\documentclass[trsc, nonblindrev]{informs3} 
\usepackage{enumitem}
\DoubleSpacedXI

\usepackage{natbib}
\usepackage{algorithm}
\usepackage{algpseudocode}
\usepackage{longtable}
\usepackage{subcaption}
\usepackage{graphicx}

\usepackage{lineno}

 \bibpunct[, ]{(}{)}{,}{a}{}{,}%
 \def\bibfont{\small}%
\TheoremsNumberedThrough     

\EquationsNumberedThrough    

\begin{document}


\RUNAUTHOR{Huang and Shen}

\RUNTITLE{Integrated Assignment and Path-Finding for Robotic Sorting}

\TITLE{Flow-Based Integrated Assignment and Path-Finding for Mobile Robot Sorting Systems}

\ARTICLEAUTHORS{%
\AUTHOR{Yiduo Huang}
\AFF{Department of Civil and Environmental Engineering, University of California Berkeley, Berkeley, CA 94720,
\EMAIL{yiduo\_huang@berkeley.edu}
}

\AUTHOR{Zuojun Shen}
\AFF{Department of Industrial Engineering and Operations Research, University of California Berkeley, Berkeley, CA 94720, \EMAIL{maxshen@berkeley.edu}, Corresponding author}
} 

\ABSTRACT{%
Express companies are deploying more robotic sorting systems, where mobile robots are used to sort incoming parcels by destination. In this study, we propose an integrated assignment and path-finding method for robots in such sorting systems. The method has two parts: offline and online. In the offline part, we represent the system as a traffic flow network, develop an approximate delay function using stochastic models, and solve the min-cost network flow problem. In the online part, robots are guided through the system according to the calculated optimal flow split probability. The online calculation of the method is decentralized and has linear complexity. Our method outperforms fast multi-agent path planning algorithms like prioritized planning because such algorithms lead to stochastic user equilibrium traffic assignment. In contrast, our method gives the approximated system-optimal traffic assignment. According to our simulations, our method can achieve 10\%--20\% higher throughput than zoning or random assignment. We also show that our method is robust even if the initial demand estimation is inaccurate.
}%


\KEYWORDS{automated warehouse, assignment, multi-robot path-finding, network flow, robotic sorting}

\maketitle

%


\section{Introduction}
\label{s:1}

The rise of e-commerce and online shopping has driven significant growth in the express delivery industry, as consumers expect their purchases to be delivered quickly and efficiently. E-commerce sales have increased 43\% in 2020 \citep{Brewster2022}, while each online order results in at least one parcel being handled. In addition, promotions and shopping festivals create large fluctuations in demand. To deal with the large demand with high uncertainty, express companies are looking for a sorting solution that is efficient enough to deal with the large parcel volume and flexible enough to handle the rapid changes in demand. A new solution is to use mobile robots to sort and transport parcels on a grid-based facility, called a robotic sorting system (RSS) \citep{Zou2021}.

In an RSS, robots move between pick-up workstations and drop-off points to transport parcels on grid roads. Incoming parcels are assigned to different workstations and loaded on empty mobile robots. Each robot can carry at most one parcel at a time. A loaded robot travels to the drop-off point corresponding to a delivery destination and drops the parcel. After dropping off, the robot becomes empty again and can be assigned to a workstation for the next parcel-handling job. Some industrial examples include tSort by Tomkins Robotics \citep{TompkinsRobotics},  Xanthus and Pegasus robots by Amazon \citep{Ames2019}, and Deppon Express \citep{Xu2022}. To see how such a system works in action, readers may refer to online videos \url{https://www.youtube.com/watch?v=4MH7LSLK8Dk&ab_channel=AmazonNews} or \url{https://www.youtube.com/watch?v=EbLDXsEPHS8&ab_channel=TompkinsSolutions}. The system is grid-based, since robots can only travel on a virtual grid of non-overlapping cells. There can be at most one robot in one cell at a time, and the robot can rotate in position in one cell. An RSS is flexible in layout and capacity: conveyor belts are unnecessary between workstations and drop-off points so that the facility can be in any shape. Expanding capacity or scale requires adding more robots, workstations, or drop-off points.

When working with an RSS, the controller must decide on two operation problems: \textit{assignment} and \textit{path-finding}. The assignment problem consists of two sub-problems: (1) parcel-to-workstation assignment and (2) robot-to-workstation assignment (also called \textit{dispatching} in the literature; see, e.g., \citet{Fransen2020}). When a new parcel arrives, it needs to be assigned to one of the workstations to be handled; when a robot drops off a parcel, it needs to be assigned to one of the workstations to be loaded. The path-finding problem is to give each robot a sequence of cells (a path) starting from its current location to the destination so that the robot can complete its job without collision. In most current real-world systems, these problems are treated separately, resulting in suboptimal operations. Our study aims to solve the assignment problem and the path-finding problem to achieve a larger throughput capacity simultaneously. We use \textit{throughput} as the performance measure, where \textit{throughput} is defined as the maximum number of parcels the RSS can handle in one hour.

The trade-off between system optimality and computation times is the major difficulty in multi-agent path-finding (MAPF). For a small system with a few robots and a finite operating time horizon, it is possible to coordinate all robots and formulate centralized mixed integer program models to find the optimal collision-free scheduling, routing, and assignment results. However, the multi-robot path-finding problem is NP-hard \citep{Yu2013}, making it impossible to find the global optimal solution for large systems. On the other hand, using the individual shortest paths for each robot without coordination results in congestion and longer waiting times. Classic multi-robot path-finding algorithms such as prioritized planning (cooperative A* in \citep{Silver2005}) try to find a real-time conflict-free solution. Such algorithms start their search from the individual shortest path and try to alter the individual paths to avoid conflicts. However, such methods can only achieve stochastic user equilibrium instead of the system optimum because they greedily try to minimize the deviation from each robot's shortest path instead of minimizing the total cost. Even if their result is conflict-free, it is still not system-optimal. We will show in Section~\ref{s:4.3} that a fast MAPF algorithm, such as prioritized planning, leads to a stochastic user equilibrium.

In this paper, we consider the steady state of a grid-based robotic sorting system, we formulate integrated robot-to-workstation, parcel-to-workstation assignment, and path-finding problems as a multi-commodity min-cost network flow problem, and we find the system-optimal flow solution using the Frank--Wolfe algorithm. The main idea of the operation is to find a real-time assignment and path-finding strategy that can create a traffic flow that is as close to the optimal flow as possible. We assign robots and parcels according to the flow split proportion on corresponding nodes. The system-optimal multi-robot paths can be found by recovering the optimal path flow and assigning robots to paths according to flow intensity. Our contributions can be summarized as follows:
\begin{itemize}
    \item A real-time integrated assignment and path-finding algorithm for RSSs.
    \item Introduction of system-optimal network flow to the grid-based mobile robot systems.
    \item A flow-delay model for grid-based mobile robot systems.
\end{itemize}

We describe related research in Section~\ref{s:2} and introduce our approach in Sections~\ref{s:3} and~\ref{s:4}. Our approach can be divided into an offline part and an online part. The offline part should be run before operating the system and only needs to be run once. There are four steps in the offline part:
\begin{enumerate}
    \item Estimate the desired throughput of the system. The manager needs to give a desired throughput level to solve the problem. Later (Section~\ref{s:5.2}), we  show that our solution is robust to this estimation and therefore does not need to be very accurate.
    \item Represent the system as a directed graph (Section~\ref{s:3}).
    \item Solve the min-cost network flow problem and obtain the optimal link flow distribution (Section~\ref{s:4.1}).
    \item Decompose link flow to path flow (Section~\ref{s:4.2.1}).
\end{enumerate}
The offline procedures give us an optimal path-flow distribution. When operating the system online, we can assign paths to robots and parcels according to the optimal path-flow intensity (Section~\ref{s:4.2.2}). We show in Section~\ref{s:5.1} that our method outperforms random assignment and the optimal-zoning method.

\section{Related Work}
\label{s:2}
To operate an RSS, the operator must solve the assignment and  path-finding problems. Although there have been only a few papers specifically focusing on RSS assignment, since this is a relatively new application,  RSSs have  similar layouts and operating rules to robotic mobile fulfillment systems (RMFSs) \citep{Azadeh2019}, which have been widely addressed in the literature. A comprehensive review of the planning and control of mobile robot systems for intralogistics can be found in \citet{FRAGAPANE2021405}.

There are two streams of research to solve the assignment problem for RSSs/RMFSs. One stream focuses on static-state analysis, using queuing networks to evaluate  assignment strategies or construct their objective functions. \citet{Lamballais2017} built a semi-open queuing network and proposed the use of a zone-based rack-to-storage assignment strategy for an RMFS. They extended their model to incorporate order replenishment and inventory management \citep{Lamballais2019} and compared different product-to-rack assignment rules. They found that spreading stock-keeping units (SKUs) among racks will improve the throughput. \citet{Roy2019} compared different robot-to-workstation assignment rules using closed queuing networks. They found that a pooled robot system outperforms a dedicated robot strategy, and assigning robots to the least congested zone will improve the performance of multi-zone systems. For RSS, \citet{Zou2021} developed a closed queuing network model to predict system throughput. \citet{Xu2022} optimized  parcel-to-workstation assignment as an integer programming problem, with the throughput estimated using an open queuing network.

Another stream of research focuses on a transient-state system with a finite time horizon. In this approach,  the assignment problem is formulated as mixed integer programming (MIP), with binary variables to indicate if a certain robot/rock/order is assigned to one workstation at a certain time. \citet{Boysen2017} optimized the robot visiting sequence and order fulfillment sequence to minimize total robot visits in an RMFS. \citet{Weidinger2018} optimized rack-to-storage position to minimize the total travel distance. They also proposed  ``shortest-path storage'' assignment rules, which can achieve similar performance to the optimal solution of their MIP. \citet{Wang2021} studied rack-to-workstation assignment optimization under human picker working state uncertainty using stochastic dynamic programming.
 
There have been numerous studies of  MAPF problems for mobile robots or automated guided vehicles (AGVs) \citep[for reviews, see][]{LeAnh2006,Yan2013,Qiu2002,DeRyck2020}. According to the review by \cite{cao2012overview},  approaches to these problems can be divided into two categories: centralized and decentralized. With centralized approaches \citep[e.g.,][]{Zelinsky1992,Silver2005,Barer,Sharon2015,Yu2016}, a central server controls all the robots in real-time. In small-scale systems, centralized algorithms can find the exact optimal collision-free solution. However, multi-robot path-finding has been proven to be NP-hard \citep{Yu2013}. The above-mentioned studies adopted different heuristics to find near-optimal solutions. Even with the heuristics, however, the dimension of the variable space can be very high, and such problems become unsolvable in real time.

Decentralized approaches scale well with the number of robots and the size of the roadmap. With a decentralized approach, each robot decides its own path and resolves conflicts using only local information. Decentralized approaches (see, e.g., \citet{yang2008multi} and the literature reviewed by \citet{zhang2013} and by \citet{cao2012overview}) are much faster and can be used by large systems in real time, but sometimes they cannot avoid gridlocks, and the solution obtained is far from optimal.

Using traffic flow to help avoid congested areas is a new decentralized method for large-scale systems for multi-robot problems. \cite{Fransen2020} used a graph representation of the system where the vertex weight is updated according to the real-time robot congestion. \citet{digani2014hierarchical,digani2015ensemble} and \citet{digani2016} developed hierarchical strategies for multi-AGVs in which  the system roadmap is partitioned into sectors, and the central planner optimizes section-level paths for robots to avoid congestion while a decentralized strategy is used to coordinate robots within each sector. \citet{digani2016} developed a Markov chain model to represent traffic evolution.

\section{System Description}
\label{s:3}
\subsection{Grid-Based Mobile Robot Sorting System and Assumptions}\label{s:3.1}
\begin{figure}[ht]
    \centering
    \includegraphics[width=13cm]{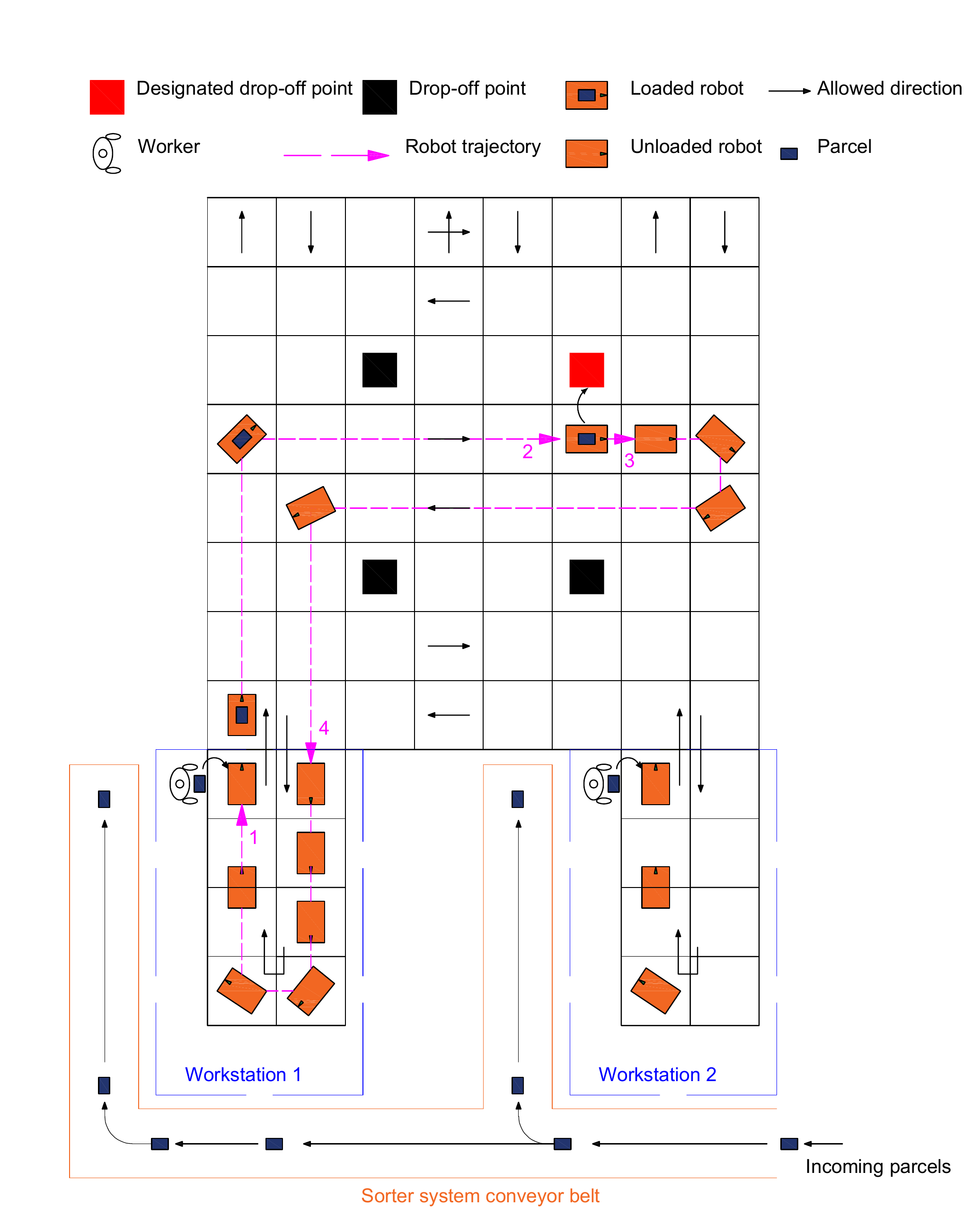}
    \caption{Layout of an Example RSS}
    \label{fig layout}
\end{figure}
The layout of the RSS studied in this paper is shown in Figure~\ref{fig layout}. The goal of the system is to handle parcels labeled with different delivery destinations and transport them to the correct drop-off points using mobile robots. Systems with similar layouts can be found in Amazon \citep{Ames2019} and Deppon Express \citep{Xu2022}. The robot moving area consists of non-overlapping square cells. Each cell can accommodate at most one robot at a time. There are two basic movements: moving from one cell to another and making a 90-degree turn. A robot can move from cell $i$ to $j$ in time $T_1$ if $j$ is next to $i$, and the robot in $i$ is heading to $j$. To make a turn in cell $i$, a robot must stop at cell $i$ and then rotate 90 degrees. We assume that it takes $T_2$ to complete a 90-degree turn.

There are some special areas in the system. Some cells are designated as drop-off points, where we put a hole in the ground, leading to a roll container installed on the lower floor. Each drop-off point corresponds to one delivery destination (represented by solid black or red squares in Figure~\ref{fig layout}). Robots can stop at the cell next to the drop-off point to drop off a parcel, but they are not allowed on these cells. We assume that dropping the parcel takes $T_\mathrm{drop}$. Some cells are designated as workstation areas (represented by the blue dashed rectangle in Figure~\ref{fig layout}), where empty robots can form queues and be loaded by human workers. The parcel loading time $T_\mathrm{load}$ can be deterministic or random.

As well as the grid of cells, there is a sorter system that can assign incoming parcels to workstations. The sorter system consists of conveyor belts at the parameter of the facility, connecting workstations to the facility parcel entrance (see Figure~\ref{fig layout}). When a new parcel arrives, the sorter system will transport it to one of the workstations. At the workstation, a human worker will load the parcel onto an empty robot, and the robot will carry the parcel to its destination drop-off point. 

The movement trajectory of a robot can be summarized in four steps (dashed purple arrows in Figure~\ref{fig layout}):
\begin{enumerate}
    \item loading at a workstation;
    \item carrying the parcel to its designated drop-off point;
    \item unloading the item at the cell next to the drop-off point;
    \item going to one of the workstations and waiting for the next parcel.
\end{enumerate}
We make the following assumptions about the system:
\begin{enumerate}
    \item The cells are unidirectional. One cell does not allow two opposite directions of movement. For example, if robots can move from south to north on cell $i$, then no robots can move from north to south on this cell (see ``allowed direction'' in Figure~\ref{fig layout}). This is a common design in real-world RSSs, described in \citet{Ames2019,TompkinsRobotics}, and \citet{Xu2022}, and can help avoid robot conflict.
    \item We ignore the acceleration/deceleration time loss and assume a constant movement speed when moving from cell to cell. Specifically, we assume it takes $T_1$ to move through one cell and $T_2$ to make a 90-degree turn.
\end{enumerate}

If there are $n_W$ workstations, $n_D$ drop-off points, and $N$ cells that are not in workstations or drop-off points, then we let $\{W_1,\dots,W_{n_W}\}$ be the set of workstations,  $\{D_1,\dots,D_{n_D}\}$  the set of drop-off points, and $\{C_1,\dots,C_N\}$  the set of cells that are neither workstation cells nor drop-off points. In the example in Figure~\ref{fig layout}, we have $n_W=2$, $n_D=4$, and $N=8\times8-4=60$.

\subsection{RSS Operation as a Multi-Commodity Traffic Flow Network}
\begin{figure}[ht]
    \centering
    \includegraphics[width=13cm]{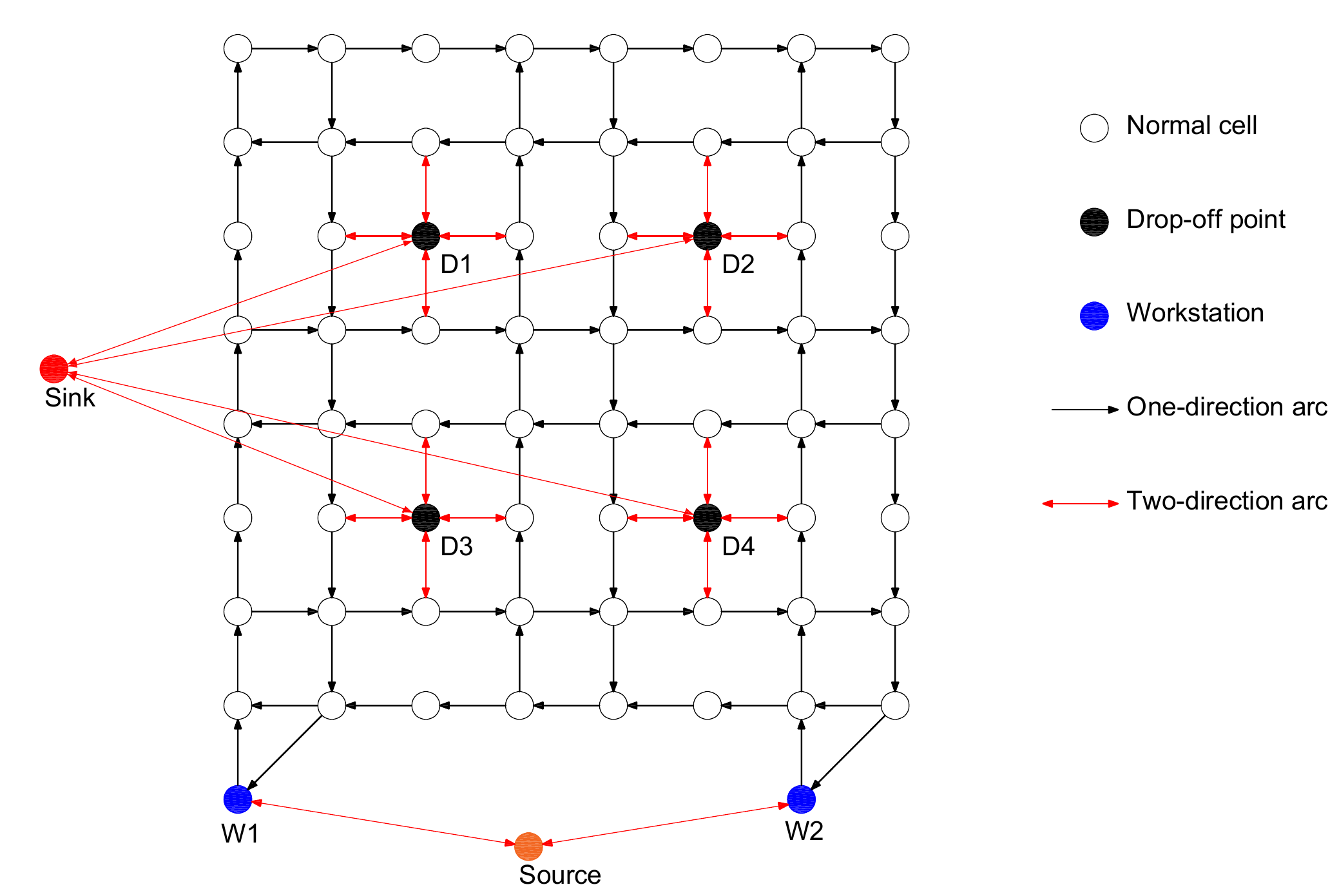}
    \caption{Flow Network Representation of the Example RSS}
    \label{fig network}
\end{figure}
Consider the RSS described in Section~\ref{s:3.1}. If the system runs smoothly without any deadlock or queue accumulation, and the parcel arrival rate equals the parcel drop-off rate, then the system is in a steady state. Under this steady state, the operation of the RSS can be characterized using a flow network. Our assignment and path-finding algorithms are based on the analysis of this network.

Consider the directed graph $\mathcal{G}=(\mathcal{V},\mathcal{A})$. The flow network representation of the example in Section~\ref{s:3.1} is shown in Figure~\ref{fig network}. Each cell, workstation, or drop-off point corresponds to one vertex in $\mathcal{V}$. $\mathcal{V}$ also include a \textit{source node} $S$ and a \textit{sink node} $T$, representing the system parcel entrance and exit. Therefore, $\mathcal{V}:=\{W_1,\dots,W_{n_W}\}\cup\{D_1,\dots,D_{n_D}\}\cup\{C_1,\dots,C_N\}\cup\{S,T\}$. 

If a robot can move from $C_i$ to $C_j$ in $T_1$, we add an arc $(C_i,C_j)$ to the set $\mathcal{A}$, representing the possible movement. Workstations are connected to normal cells that serve as their entrance or exit. Drop-off points are connected using two-direction arcs to the cells where a robot can release the parcel. In the example in Figure~\ref{fig layout}, robots can release parcels from the cell on the north, south, east, and west of the drop-off points, corresponding to the four red two-direction arcs in Figure~\ref{fig network}. In addition, we connect workstations and the source node using two-direction arcs, representing the sorter system, and we connect all the drop-off points with the sink node using two-direction arcs. 

There are two types of flow: forward flow and backward flow. Forward flow originates from the source node $S$ and ends at the sink node $T$, representing the flow of parcels through the system. Backward flow originates from $T$ and ends at $S$, representing the flow of empty robots returning to workstations. Let the set of acyclic routes from $S$ to $T$ be $\mathcal{R}_F$, and the set of routes from $T$ to $S$ be $\mathcal{R}_B$. Let the flow intensity on $r\in \mathcal{R}_F$ be $f_r^F$ and the flow intensity on $r\in \mathcal{R}_B$ be $f_r^B$. Note that by adding the source and sink node, the parcel-to-workstation and robot-to-workstation assignment information is included in the path information since the workstations and drop-off points are special nodes (the second or the last but one node on the path).

In the steady state, if we know all the $f_r^F$ and $f_r^B$, we know how the system operates. The physical meaning of $f_r^F$ is that for a path $r=(S,W_i,C_m,\dots,C_n,D_j,T)$, there are $f_r^F$ parcels loaded onto robots at workstation $W_i$ and dropped at $D_j$ per hour, while the robot carrying these parcels will follow the path $(C_m,\dots,C_n) \subset r$. For $f_r^B$, $r=(T,D_j,C_n,\dots,C_m,W_i,S)$, there are $f_r^B$ robots returning from $D_j$ to $W_i$ following the path $(C_n,\dots,C_m)$ per hour. The cost on route $r$ (machine time spent per hour) is
\begin{equation}
    RC_r(\mathbf{f}) = \sum_{(i,j)\text{ on path } r}c_{ij}(\mathbf{f})+ \sum_{(i,j),(j,k)\in r}T_2 \delta^{turn}_{i,j,k},
    \label{eq RC}
\end{equation}
where $\mathbf{f}:=[\mathbf{f}^F,\mathbf{f}^D]=[[f_r^F],[f_r^B]]$ is the vector of all path-flow variables, and $c_{ij}(\mathbf{f})$ is the expected travel time from cell $C_i$ to $C_j$.  $\delta^{turn}_{i,j,k}=1$ if the robot going from cell $C_i$ to $C_j$ then to $C_k$ needs to make a 90-degree turn at cell $C_j$; otherwise, $\delta^{turn}_{i,j,k}=0$. The first term in Equation~(\ref{eq RC}) is the arc travel time incurred by traveling from cell to cell and waiting when blocked. The second term is the extra turning time because it takes $T_2$ to make a 90-degree turn. The total system cost can be estimated as
\begin{equation}
    TC(\mathbf{f}) =\sum_{r\in\mathcal{R}_F} {f_r^F RC_r(\mathbf{f})} +
    \sum_{r\in\mathcal{R}_B} {f_r^B RC_r(\mathbf{f})}.
    \label{eq TC}
\end{equation}
The arc flow can be uniquely determined by the path flow. Letting $v_{ij}$ be the flow on the arc $(C_i,C_j)$, we have
\begin{equation}
    v_{ij} = \sum_{r\in \mathcal{R}_F} f_r^F\delta_{ijr} + \sum_{r\in \mathcal{R}_B} f_r^B\delta_{ijr},
    \label{eq vij}
\end{equation}
where $\delta_{ijr}=1$ if $(i,j)$ is on path $r$ and $\delta_{ijr}=0$ otherwise.

\section{Flow-Based Integrated Assignment and Path-Finding}
\label{s:4}
\subsection{Finding the System Optimal Flow}\label{s:4.1}
The analysis of in this subsection is based on the following assumptions
\begin{enumerate}
    \item \textbf{Sufficient robots} There are enough empty robots waiting in the buffer areas of workstations that the waiting time for robots at the loading station can be ignored.
    \item \textbf{Steady state} We only consider the system under a steady state. We assume that there is no gridlock and that the workstation has enough capacity to handle incoming parcels.
    \item \textbf{Poisson flow} We assume that the flow on the network is memoryless. In a real-world system, robots will circulate in the system, and any cell/workstation in the system can be viewed as a server with a feedback flow. Similar to the Poisson traffic flow in feedback queues \cite{Pekoz2002}, the traffic flow is asymptotic Poisson if the system is not saturated (i.e., when there are no unsolvable gridlocks or excessive queues).
    \item \textbf{First-come-first served with ties} We assume a robot will only check one cell ahead when executing the route. If the cell ahead is already occupied, the robot stops and waits. If two robots request the same cell simultaneously, we assume that each has the same chance to be granted access first.
\end{enumerate}
We need these assumptions to simplify the analysis of the flow network. We will solve the optimal network flow with these assumptions to minimize the total cost equation~\eqref{eq TC}. Although some assumptions may not be realistic, the optimal flow is not used directly in the online routing and assignment. It only provides a rough estimation of the desired traffic distribution, which is enough to improve the system's performance, as shown in our numerical examples.

\subsubsection{Cell Delay Function.}
\begin{figure}[ht]
    \centering
    \includegraphics[width=4cm]{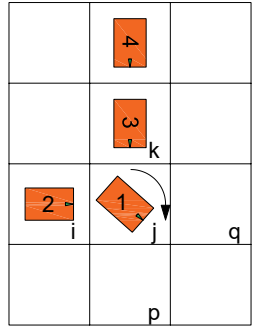}
    \caption{A Typical Blocking}
    \label{fig blocking}
\end{figure}
The major difficulty in estimating the cost in Equations~\eqref{eq TC} and~\eqref{eq RC} is the stochastic travel time $c_{ij}(\mathbf{f})$. Consider cells $C_i$, $C_j$, $C_k$, $C_p$, and $C_q$ in Figure~\ref{fig blocking}, and take a snapshot when robot 2 just stops in $C_i$. When a robot moves from cell $C_i$ to $C_j$ ($C_i$ and $C_j$ are next to each other), if other robots do not block $C_j$, then it will take $T_1$ to complete the movement. If cell $C_j$ is blocked, the robot in cell $C_i$ must wait until $C_j$ is cleared. Therefore, the travel time can be decomposed into two parts:
\begin{equation}
    c_{ij}(\mathbf{f}) = T_1 + \mathbb{E}[S_{ij}]
    \label{eq cij}
\end{equation}
$c_{ij}(\mathbf{f})$ includes a deterministic part $T_1$, which is the time required for one robot to move one cell length, and a stochastic part $\mathbb{E}[S_{ij}]$, where $S_{ij}$ is the random waiting time that robot in cell $C_i$ needs to wait until cell $C_j$ is cleared, $S_{ij}=0$ if  cell $C_j$ is empty. Let $C_k$ be the other cell where robots can enter $C_i$, $B_j$ be the time that a robot will spend in $C_j$ (the robot service time), $R_{j}$ be the random clearance time (the unfinished part of $B_j$ when a new robot arrives) in $C_i$, and $I_{k,j}\in\{0,1\}$ be the number of robots on $C_k$ heading to $C_j$. For example, in Figure~\ref{fig blocking}, $S_{ij}$ is the time for robot 2 to wait until $C_j$ is empty. $R_{j}$ is the time for robot 1 to complete the turning operation and leave $C_j$. $I_{k,j}=1$, since robot 3 is in $C_k$. $B_j$ is the total time that robot 1 spent in $C_j$. Then, the expected waiting time for a robot in $C_i$ is
\begin{equation}
    \mathbb{E}[S_{ij}] = \mathbb{E}[R_{j}] + 0.5\mathbb{P}(I_{k,j}=1)\mathbb{E}[B_{j}]
    \label{eq Sij}
\end{equation}
The first term on the right-hand side is the expected time for robot 1 to complete. The second term is the expected time for extra waiting if the robot from $C_k$ has higher priority. For example, if robot 2 has higher priority than robot 3, then $S_{ij}=R_j$. Otherwise, robot 2 needs to wait until robot 3 is done with $C_j$, and $S_{ij}=R_j+B_j$. According to our assumption of steady Poisson arrival, the probability that 2 arrives earlier than 3 and 3 earlier than 2 are equal, so we have $0.5$ as the probability.

Since $C_j$ has two possible states: blocked or not-blocked, with continuous time horizon, the blocking-unblocking process can be modeled as an alternating renewal process, where the blocking time in one cycle is $B_j$, and the cycle length (i.e. arrival interval) is $(v_{ij}+v_{jk})^{-1}$. Let the total flow on $C_j$ be $v_{j}:=v_{ij}+v_{kj}$. Using the renewal theorem and our assumption of Poisson flow, we have
\begin{equation}
    \mathbb{E}[R_j] = \frac{v_j}{2} \mathbb{E}[B_j^2]
    \label{eq ERj}
\end{equation}
Consider the alternating renewal process in $C_k$, using the renewal theorem:
\begin{equation}
    \mathbb{P}(I_{k,j}=1) = v_{kj}\mathbb{E}[B_k]
    \label{eq Nkj}
\end{equation}
Plugging Equations~\eqref{eq ERj} and~\eqref{eq Nkj} into~\ref{eq Sij}, we have
\begin{equation}
     \mathbb{E}[S_{ij}] =  \frac{v_j}{2} \mathbb{E}[B_j^2]+ 
     \frac{v_{kj}}{2}\mathbb{E}[B_k]\mathbb{E}[B_{j}]
     \label{eq Sij as BjBk}
\end{equation}
Note that there are two terms on the right-hand side of Equation~\eqref{eq Sij as BjBk}: the first is the waiting time caused by the remaining work on $C_j$, and the second is the waiting time caused by competing robots coming from cell $C_k$.

\subsubsection{Approximated Cell Delay Function.}
\label{s:4.1.2}
The distribution of $B_j$ depends on the blocking status of the downstream cells $C_p$ and $C_q$. Under light traffic, $C_p$ and $C_q$ are usually empty, and robot 1 in the example cell $C_j$ does not need to stop and wait. For cell $j$, let the through traffic without dropping flow be $v_j^{(1)}$, the turning traffic flow be $v_{j}^{(2)}$, and the flow of dropping-off robots be $v_{j}^{(3)}$ ($v_{j} = v_j^{(1)}+v_j^{(2)}+v_j^{(3)}$), i.e.,
\begin{equation}
    v_j^{(l)}:=\sum_{r\in \mathcal{R}_F} f_r^F\delta_{j,r}^{(l)} + \sum_{r\in \mathcal{R}_B} f_r^B\delta_{j,r}^{(l)}, \qquad l=1,2,3,
    \label{eq vjl}
\end{equation}
where $\delta_{j,r}^{(1)}=1$ if robots on path $r$ will go through cell $C_j$,  $\delta_{j,r}^{(2)}=1$ if robots on path $r$ will make a turn on cell $C_j$, $\delta_{j,r}^{(3)}=1$ if robots on path $r$ will drop parcels on cell $C_j$, and $\delta_{j,r}^{(l)}=0$ otherwise.

Let $G_j$ be the service time on $C_j$ when the downstream cells $C_p$ and $C_q$ are empty. $G_j$ depends entirely on whether the robot will make a turn or drop a parcel. If the robot does neither of these, then  $G_j=2T_1$. If the robot makes a turn, then $G_j=2T_1+T_2$. If the robot drops a parcel, $G_j=2T_1+T_\mathrm{drop}$. Therefore, $G_j$ is a categorical distributed random variable, with $\mathbb{P}(G_j=2T_1) = v_j^{(1)}/v_j$, $\mathbb{P}(G_j=2T_1+T_2) = v_j^{(2)}/v_j$, and $\mathbb{P}(G_j=2T_1+T_\mathrm{drop}) = v_j^{(3)}/v_j$.

If we assume that all ``downstream'' cells $C_p$ and $C_q$ are empty, $B_j=G_j$, then we have the following expression for delay using Equation~\eqref{eq Sij approx} and the distribution of $G_j$:
\begin{align}
    \mathbb{E}[S_{ij}] \approx {}& \frac{v_j}{2} \mathbb{E}[G_j^2]+ 
     \frac{v_{kj}}{2}\mathbb{E}[G_k]\mathbb{E}[G_{j}]\\
    =&\frac{1}{2}[4v_j^{(1)}T_1^2+v_j^{(2)}(2T_1+T_2)^2+v_j^{(3)}(2T_1+T_\mathrm{drop})^2]\nonumber\\
    &+\frac{1}{2v_k}[2v_j^{(1)}T_1+v_j^{(2)}(2T_1+T_2)+v_j^{(3)}(2T_1+T_\mathrm{drop})]\nonumber\\
    &\hspace{30pt}\times [2v_k^{(1)}T_1+v_k^{(2)}(2T_1+T_2)+v_k^{(3)}(2T_1+T_\mathrm{drop})]
    \label{eq Sij approx}
\end{align}
Let the max flow through any cell be $v_{max}:=\max_{C_k \in {C_1,...,C_N}}(v_k)$We then show that the approximation error is in $O(v_{max}R^2)$.
\begin{theorem}
    \[\frac{v_j}{2} \mathbb{E}[G_j^2]+ 
     \frac{v_{kj}}{2}\mathbb{E}[G_k]\mathbb{E}[G_{j}]
     \leq \mathbb{E}[S_{ij}] \leq  
     \frac{v_j}{2} \mathbb{E}[G_j^2]+ 
     \frac{v_{kj}}{2}\mathbb{E}[G_k]\mathbb{E}[G_{j}] + \max{(v_q,v_p)} R^2(C_G)^2\]
\label{theorem Sij error}
\end{theorem}

\proof{Proof.}
Note this is the approximated delay function based on the assumption that $C_p$ and $C_q$ are empty. With our assumption on Poisson arrival, the probability that when a robot arrives in $C_j$, downstream cell $C_p$ or $C_q$ are blocked is
\begin{equation}
    \mathbb{P}(B_j\neq G_j) =\frac{1}{v_j}(v_{jq}v_{q}\mathbb{E}[B_q]+v_{jp}v_{p}\mathbb{E}[B_p])
\end{equation}

Note that $B_k$ is bounded for any cell $C_k$. If there are $R$ robots, the worst case is that one robot needs to wait for the other $R-1$ robots. The time that one robot used one cell is bounded by $C_G :=\max{(T_2+2T_1, T_{drop}+2T_1)}$
\begin{equation}
    B_k \leq R C_G
\end{equation}
Therefore, the probability that our approximation is not correct is bounded,
\begin{equation}
    \mathbb{P}(B_j\neq G_j) \leq \max{(v_q,v_p)} RC_G
\end{equation}
In addition, $B_j\geq G_j$ since $B_j$ includes blocking waiting time while $G_j$ does not. The estimation error bound is
\begin{equation}
     \frac{v_j}{2} \mathbb{E}[G_j^2]+ 
     \frac{v_{kj}}{2}\mathbb{E}[G_k]\mathbb{E}[G_{j}]
     \leq \mathbb{E}[S_{ij}] \leq  
     \frac{v_j}{2} \mathbb{E}[G_j^2]+ 
     \frac{v_{kj}}{2}\mathbb{E}[G_k]\mathbb{E}[G_{j}] + \max{(v_q,v_p)} R^2(C_G)^2
\end{equation}
Therefore, we can see the estimation error increases in $O(v_{max}R^2)$
\endproof
\Halmos

\subsubsection{Loading and Dropping Delay Function.}
Workstation $W_k$ needs to process $v_{\mathrm{source}, W_k}$ parcels per unit  time. We assume that the first two moments of the human worker processing time $T_\mathrm{load}$, namely, $\mathbb{E}[T_\mathrm{load}]$ and $\mathbb{E}[T_\mathrm{load}^2]$, are known. Since we assume that there are enough robots in the system and that the arrival process of parcels is Poisson, workstations can be modeled as M/G/1 queues. According to the Pollaczek--Khinchin mean formula, the expected travel time from the source node to workstation $W_k$ is the mean waiting time: 
\begin{equation}
    c_{\mathrm{source},W_k}(\mathbf{f})=\mathbb{E}[T_\mathrm{load}]+\frac{v_{\mathrm{source},W_k}\mathbb{E}[T_\mathrm{load}^2]}{2(1-v_{\mathrm{source},W_k}\mathbb{E}[T_\mathrm{load}])}.
    \label{eq cSWk}
\end{equation}
The waiting time for drop-off points is already calculated in the cell waiting time, since the queues for drop-off points form in the cells near the drop-off point and there is no dedicated waiting buffer for dropping-off robots. Therefore, the expected travel time from cell $C_i$ to drop-off point $D_k$ is (if $C_i$ is connected to $D_k$)
\begin{equation}
    c_{i,D_k}(\mathbf{f})=\mathbb{E}[T_\mathrm{drop}].
    \label{eq ciDk}
\end{equation}

\subsubsection{Solving the Min-Cost Network Flow.}
\label{s:4.1.4}
Under the steady state, the approximated optimal integrated assignment-routing problem can be formulated as
\begin{align}
    &\hspace{-46pt}\min\ TC(\mathbf{f})
    \label{eq min cost}\\
\text{s.t. } &\text{Equations~(\ref{eq RC})--(\ref{eq cij}), (\ref{eq vjl})--(\ref{eq Sij approx}), and (\ref{eq cSWk})--(\ref{eq ciDk}) hold and}\nonumber\\
    &d_{D_k}=\sum_{r\in\mathcal{R}_F: D_k\in r} f_r^F,
    \label{eq demand forward}
\\
    &d_{D_k}=\sum_{r\in\mathcal{R}_B: D_k\in r} f_r^B.
    \label{eq demand backward}
\end{align}
The constraints~(\ref{eq demand forward}) and~(\ref{eq demand backward}) state that the total forward and backward flow should satisfy the parcel demand for any drop-off points. Note that the parcel demand must be given before solving the problem. In real-world problems, since we do not know the real throughput capacity of the system, we need a rough estimate of the demand.

We can use the Frank--Wolfe method to find the solution to solve the min-cost network flow problem~(\ref{eq min cost}). At each step $\mathit{iter}$,  we consider the linearized subproblem
\begin{align}
 &\hspace{-46pt}\min_\mathbf{f}\ \label{eq linear subprob} TC(\mathbf{f}^{(\mathit{iter})})+\frac{\partial TC}{\partial{\mathbf{f}^{(\mathit{iter})}}}(\mathbf{f}-\mathbf{f}^{(\mathit{iter})})
\\
\text{s.t. }&\text{Equations~(\ref{eq RC})--(\ref{eq cij}), (\ref{eq vjl})--(\ref{eq Sij approx}), (\ref{eq cSWk})--(\ref{eq ciDk}), and (\ref{eq demand forward})--(\ref{eq demand backward}) hold}.\nonumber
\end{align}

\begin{figure}[ht]
    \centering
    \includegraphics[width=10cm]{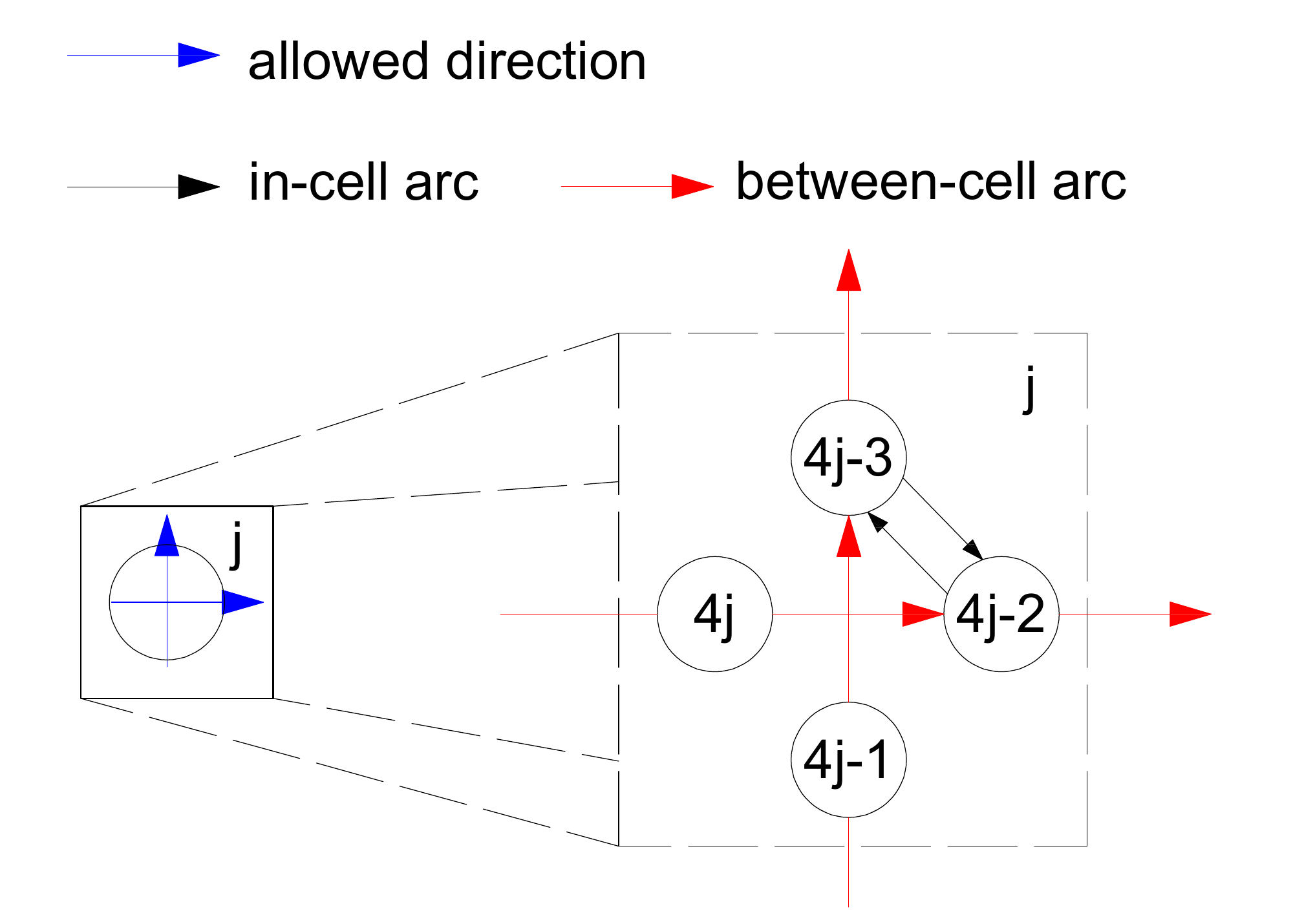}
    \caption{Cell Node Decomposition}
    \label{fig cell nodes}
\end{figure}

Each linear subproblem can be solved using a linear programming solver. The Frank--Wolfe method includes the following steps:
\begin{enumerate} [label= Step \arabic*:] 
    \item Find the free-flow shortest path for all source--workstation pairs. Assign flow $\mathbf{f}^{(\mathit{0})}$ using free-flow shortest path assignment. Set $\mathit{iter}=0$.
    \item Solve the linear subproblem (\ref{eq linear subprob}) to obtain the search direction $\mathbf{f}_\mathrm{linear}^{(\mathit{iter})}$.
    \item Do a line search from $\mathbf{f}^{(\mathit{iter})}$ to $\mathbf{f}_\mathrm{linear}^{(\mathit{iter})}$ to find the optimal objective value $TC^{(\mathit{iter})}$ and step size $\alpha^{(\mathit{iter})}$. Update the flow assignment $\mathbf{f}^{(\mathit{iter}+1)}=\mathbf{f}^{(\mathit{iter})}+\alpha^{(\mathit{iter})}(\mathbf{f}^{(\mathit{iter})}_\mathrm{linear}-\mathbf{f}^{(\mathit{iter})})$. $\mathit{iter}=\mathit{iter}+1$.
    \item  Convergence check: if $|TC_\mathrm{linear}^{(\mathit{iter})}-TC_\mathrm{linear}^{(\mathit{iter}-1)}|<\epsilon$.
\end{enumerate}
We can use the link-flow representation instead of the path flow to reduce the number of variables. Since we need to calculate the turning flow on each cell, we further decompose one cell node $C_j$ into at most four nodes and relabel the nodes: $C'_{4j}$,  $C'_{4j-1}$, $C'_{4j-2}$, $C'_{4j-3}$ (see Figure~\ref{fig cell nodes}). Each of the nodes represents one of the four headings of the cell. Note that at most two nodes in each cell are connected, since one cell only allows two directions. Using this representation, the turning flow can be calculated using the flow on in-cell arcs. We can solve the optimization problem~\eqref{eq min cost} using link-flow variables $\mathit{v}$ instead of $\mathit{f}$.

\subsection{Flow-Based Assignment and Path-Finding}
\label{s:4.2}
\subsubsection{Recovery of Path Flow.}
\label{s:4.2.1}
In Section~\ref{s:4.1.4}, we presented the path-flow formulation of the min-cost problem. Since there are exponentially many paths, we use a link-flow-based formulation with $O(|\mathcal{V}|)$  flow conservation constraints and $O(|\mathcal{V}|)$  link-flow variables. However, the link flow does not uniquely determine the path flow. In urban transportation, we must solve an entropy-maximization problem like that in \citet{BARGERA20101022} to recover the path flow. We can control all the robots in our problem, and so we do not need to maximize the entropy: any possible path flows can be used.

To recover any path flow, we use Algorithms~\ref{alg:flow decomp} and~\ref{alg:follow path} for forward and backward flows. In words, we loop over all nodes with a positive influx. Starting from each imbalanced node $i$, we do a depth-first search using the arc-flow as the weight until we reach a node $k$ with a negative influx. Then, we push flow along the path from $i$ to $k$ without changing the signs of arc flows or node influxes and obtain the residual graph. We repeat this process until all nodes are balanced. This procedure stops in finite step and can find one of the possible path-flow.

\begin{algorithm}[ht]
\caption{Flow Decomposition}\label{alg:flow decomp}
\begin{algorithmic}
\State Initialize $\mathit{\mathit{PathFlow}}[i,j]\gets 0$ for node $i$ with positive influx, $j$ with negative influx.
\For{Each node $i$ with positive influx}
\State $\mathit{Influx}_i \gets \sum_{j: (i,j)\in\mathcal{A}} v_{ij} - \sum_{j: (j,i)\in\mathcal{A}} v_{ij}$
\While{$\mathit{Influx}_i > 0$}
    \State $\mathit{Path}_{i,k}$, $k \gets$ Follow path ($i$) \Comment{See Algorithm 2}
    \State $\mathit{AssignedFlow}_{i,k} \gets \min\{\mathit{Influx}_i, -\mathit{Influx}_k, \min_{(p,q)\text{ on }\mathit{Path}_{i,k}}{v_{p,q}}\}$
    \State $\mathit{PathFlow}[i,k] \gets \mathit{PathFlow}[i,k] + \mathit{AssignedFlow}_{i,k}$
    \For{Each arc $(p,q)$ on $\mathit{Path}_{i,k}$}
        \State $v_{p,q}=v_{p,q}-\mathit{AssignedFlow}_{i,k}$ \Comment{Push flow along the path}
    \EndFor
    \State $\mathit{Influx}_i= \mathit{Influx}_i-\mathit{AssignedFlow}_{i,k}$ 
    \State $\mathit{Influx}_k= \mathit{Influx}_k+\mathit{AssignedFlow}_{i,k}$ 
\EndWhile
\EndFor
\State \Return $\mathit{PathFlow}[i,j]$
\end{algorithmic}
\end{algorithm}

\begin{algorithm}[ht]
\caption{Following Path}\label{alg:follow path}
\begin{algorithmic}
\Require start node $i$
\State $\mathit{DestSet} \gets $ nodes with negative influx
\State $\mathit{CurrentNode} \gets i$
\State $\mathit{Path} \gets [i]$
\While{$\mathit{CurrentNode}$ not in $\mathit{DestSet}$}
    \State $NextNode \gets k$: $k$ random choice from $\mathit{OutNeighbor}(\mathit{CurrentNode})$ with probability weight $v_{\mathit{CurrentNode}, k}$
    \State $\mathit{CurrentNode} \gets \mathit{NextNode}$
    \State $\mathit{Path}$ append $\mathit{CurrentNode}$
\EndWhile
\State \Return $\mathit{Path}$, $\mathit{CurrentNode}$
\end{algorithmic}
\end{algorithm}

\begin{proposition}
The flow decomposition algorithm (Algorithm \ref{alg:flow decomp}) ends in finite steps and can return one of the path-flow representations.
\end{proposition}
\proof{Proof.}
In each push operation, we either removed one edge with a positive flow or balanced one node. Therefore, we need at most $2|\mathcal{V}|+|\mathcal{A}|$ pushes. When stopped, unconnected imbalanced nodes cannot exist unless flow conservation is violated in the link flow assignment. There cannot be any remaining links with positive flows because this will result in a cycle, the link cost function is non-decreasing with flows, and the link flow assignment is optimal. Otherwise, we can subtract this flow from the cycle for a better arc flow solution. Since the residual graph is balanced with zero flow, we obtain one of the possible path-flow representations.
\endproof
\Halmos

\subsubsection{Real-Time Path-Finding and Assignment.}
\label{s:4.2.2}
Once we have the optimal path-flow distribution, we can use the flow-split probability for every node to guide robots in real-time. Specifically, if the optimal forward and backward path flows $f_r^F$ and $f_r^B$ are known, then the probability that we assign one parcel for drop-off point $d$ to workstation $w$ following path $r$ is ${f_r^F}/{\sum_{r_1: d \text{ on } r_1}f_{r_1}^F}$. The probability that we assign one empty robot from $d$ to workstation $w$ following path $r$ is ${f_r^B}/{\sum_{r_1: d \text{ on } r_1}f_{r_1}^B}$
Since all these path flows are calculated offline with at most $2|\mathcal{V}|+|\mathcal{A}|$ path flows, this algorithm runs efficiently.

After assigning the path to one robot, the robot will follow the path. Any collision-resolving algorithms can be used. For example, the system can keep a booking table where robots claim their paths several steps in advance and stop when a collision is imminent. Other decentralized traffic control algorithms, such as that in \citet{Olmi2008}, can also be used to adjust the speed curves to avoid collisions and deadlocks locally. 

Many decentralized multi-robot algorithms are based on the shortest path and speed-curve control; see, for example, \citet{jager2001decentralized}, \citet{Olmi2008}, and \citet{Peng2005}. In such algorithms, each robot follows the shortest path while communicating with other robots and stops when possible conflict is detected. All of them can be used as our traffic controllers.

\subsection{Better Performance Compared with Priority Planning}\label{s:4.3}
We will now clarify why our method is better than our benchmark. One of the most commonly used methods for fast MAPF is prioritized planning (also called cooperative A* or CA*)\citep{Silver2005}. When a new task arrives in the system, it searches for the fastest path without interrupting existing plans. More specifically, the prioritized planning method maintains a time-expansion graph. It makes $T$ copies of the network, and every vertex represents a pair $(i,t)$, where $i$ is the node index and $t$ is the time step. $(i,t_1)$ and $(j, t_2)$ are connected only if a robot in $i$ at time $t_1$ can move to its neighbor $j$ in $t_2-t_1$. If one path is assigned to one robot, then all the nodes that the robot passes on the time-expansion graph will be blocked. When a new robot with lower priority arrives, it will search for the shortest path on the  time-expansion graph without using blocked nodes. Usually, the newest task has the lowest priority, and so the previously assigned paths will not change for computational simplicity. We can show the following.

\begin{theorem}
CA* leads to a stochastic user equilibrium under a steady state. 
\end{theorem}

\proof{Proof.} Let $\mathcal{R}$ be the set of all paths connecting one origin--destination pair. For path $r\in\mathcal{R}$, the total travel time is
\[c_r:=\alpha_r T_1 + \beta_r T_2 + \sum_{(i,j):(C_i, C_j) \in r} S_{ij},\]
where $\alpha_r$ is the number of movements from one cell to another on $r$, $\beta_r$ is the number of 90-degree turns on path $r$, and $S_{ij}$ is the random waiting time when entering from $C_i$ to $C_j$.

When planning a path using CA* for one robot, the other $R-1$ robots are already executing their paths, so the new path always has the lowest priority. This robot will choose the path with the lowest travel time. In a steady state, $S_{ij}$ is a time-independent random variable determined by the flow distribution of the other $R-1$ robots, and the probability of choosing path $r$ is
\[p_r=\mathbb{P}(c_r\leq c_p:\forall p\notin r; p,r \in \mathcal{R}).\]
Therefore, the flow distribution will be a stochastic user equilibrium.
\endproof
\Halmos

Moreover, the flow choosing probability is approximately multinomial probit in a large system with light traffic (when $v_{max}RC_G<< 1$). Under light traffic, blocking only happens when one robot blocks another, and no long queues can be observed. The waiting times $S_{ij}$ are independent with finite supports. If the system is large and the path is long, we can use the central limit theorem, and so  the random term $\sum_{(i,j):C_i, C_j \in r} S_{ij}$ is normally distributed. Therefore, $\mathbf{c}:=(c_r: r\in \mathcal{R})$ is multivariate normal:
\[p_r=\mathbb{P}(c_r\mathbf{1}-\mathbf{c} \leq 0 ).\]
Note that $c_r\mathbf{1}-\mathbf{c}$ is a linear transform of $\mathbf{c}$, and so it is also multivariate normal. $p_r$ is a cumulative joint distribution function at $0$ of a multivariate normal random vector.

Since our method seeks an approximate system optimum, whereas priority planning can only achieve stochastic user equilibrium, our method will be better if the approximation is relatively accurate. We don't need to have very accurate flow distribution or objective function to outperform CA*.

\section{Simulation Experiments}
\label{s:5}
\begin{figure}[ht]
    \centering
    \includegraphics[width=10cm]{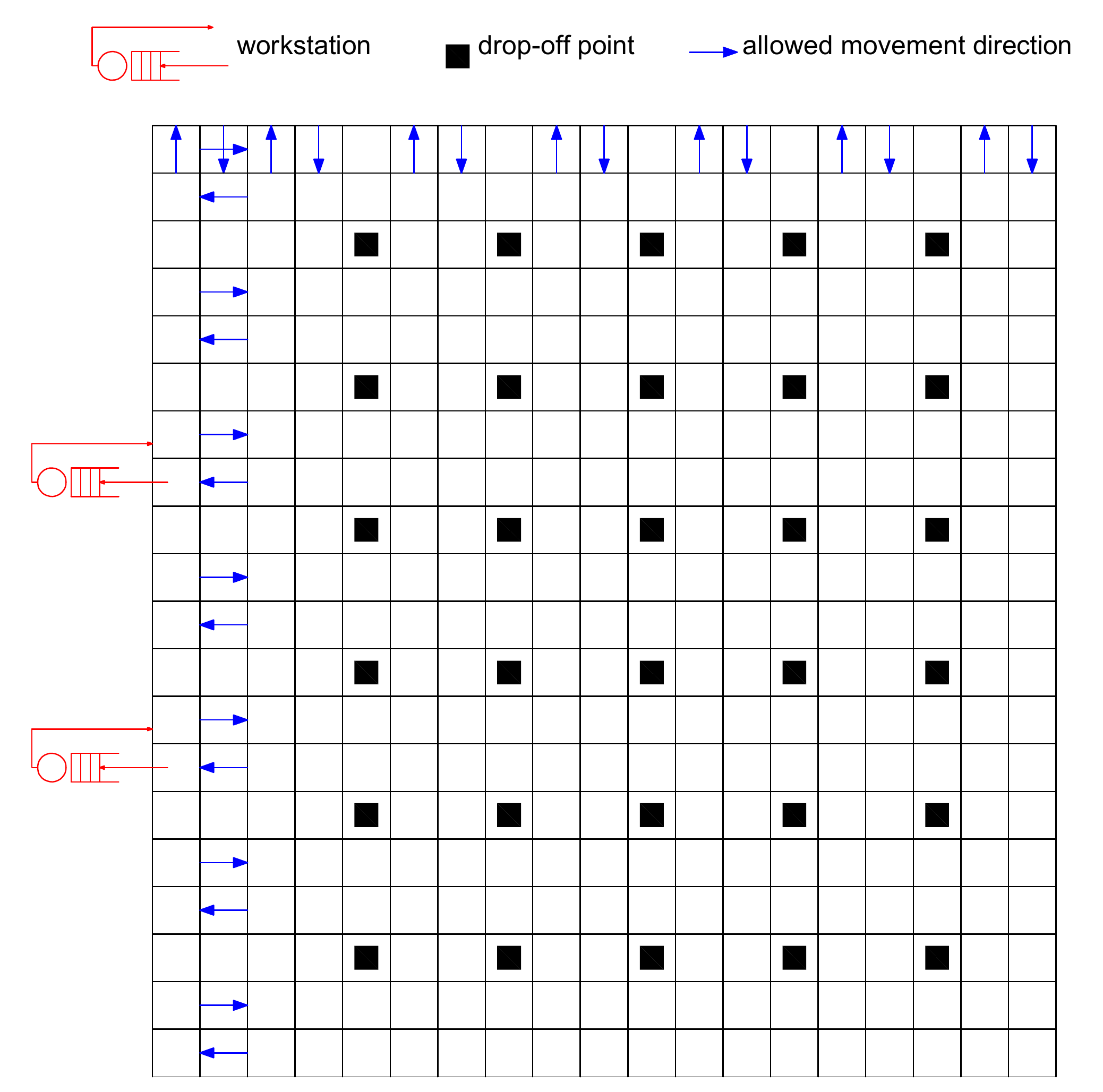}
    \caption{Simulation System Roadmap}
    \label{fig sim setting}
\end{figure}

To verify the effectiveness of our proposed method, we conducted simulations in the system shown in Figure~\ref{fig sim setting}, a small (19*20) RSS with two workstations and 30 drop-off points. We set $T_1=1$, $T_2=4$, $T_\mathrm{load}=3$, and $T_\mathrm{drop}=1$. We set constant pickup and drop-off times to simplify the calculation. We ran 3000 steps for each simulation and conducted 50 experiments for each experiment. We assumed that the probability that a parcel goes to each drop-off point is equal. We immediately assign tasks to the robot to see the system's max throughput capacity, so there's no idle robot in the system.

Our traffic control algorithm is adapted from that in \citet{jager2001decentralized}. When two robots are close to each other, they find the shortest execution path, which avoids all conflicting areas on the task completion diagram \citep[Figure 2]{jager2001decentralized}. To resolve deadlocks, we maintain a directed graph representing robot blocking \citep[Figure 3]{jager2001decentralized} to detect deadlocks and re-plan paths for robots in deadlocks using the alternative paths if a deadlock is detected. Note that this deadlock-resolving algorithm cannot guarantee deadlock-free operation. If there are too many robots in the system, deadlock is unavoidable.

\subsection{Accuracy of the Objective Function}
\begin{figure}[ht]
    \centering
    \includegraphics[width=8cm]{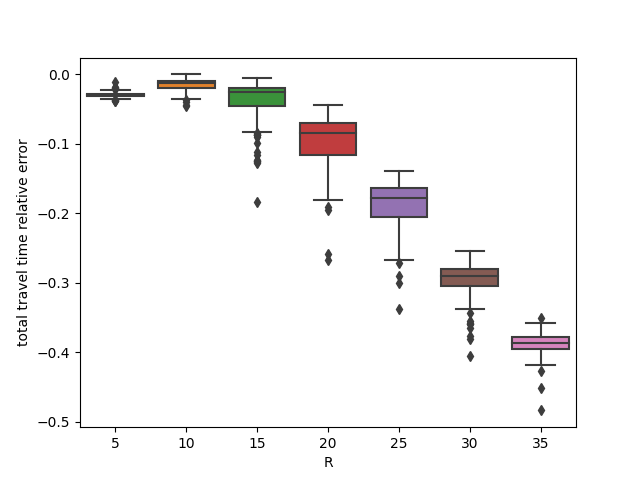}
    \caption{Relative Error in Objective Function}
    \label{fig sim relative error}
\end{figure}

Before comparing different path-finding methods, we verify our objective function \eqref{eq min cost}. We use CA* for pathfinding and random assignment rules. For each experiment setting $R$ (number of robots), we conduct 150 trials. For each trial, we run a simulation, record the simulated arc flow on each arc, and then plug them into our objective function. The relative error in the objective function is shown in Figure~\ref{fig sim relative error}.

We can see the relative error increases when there are more robots. According to Theorem \ref{theorem Sij error}, the error increase in $O(v_{max} R^2)$. Since the flow increase in $O(v_{max})$ if the system is in a steady state, the relative error should increase in $O(R^2)$. Our observation shows that relative error increases approximately in $O(R^2)$. However, we cannot draw conclusions due to the significant deviation in each experiment group. Also, our analysis didn't consider deadlocks, but deadlocks are common if there are more than 20 robots in the system. Deadlocks result in a smaller flow with a longer waiting time. In addition, during the deadlock-resolving process, the traffic flow is no longer time-independent. Our assumption on the steady system and Poisson arrival are violated, so the objective function significantly underestimated the travel time (with a relative error up to 50\%). 

Although the objective function has significant errors if there are too many robots or the flow is large. As we will show in the following sections, we can find an optimal flow distribution for a system with light traffic (i.e., our initial estimation in RHS of \eqref{eq demand forward} and \eqref{eq demand backward} is small) and use it to guide the robots in a system with more robots.

\subsection{Improving the Throughput}
\label{s:5.1}
\begin{figure}[ht]
    \centering
    \includegraphics[width=16cm]{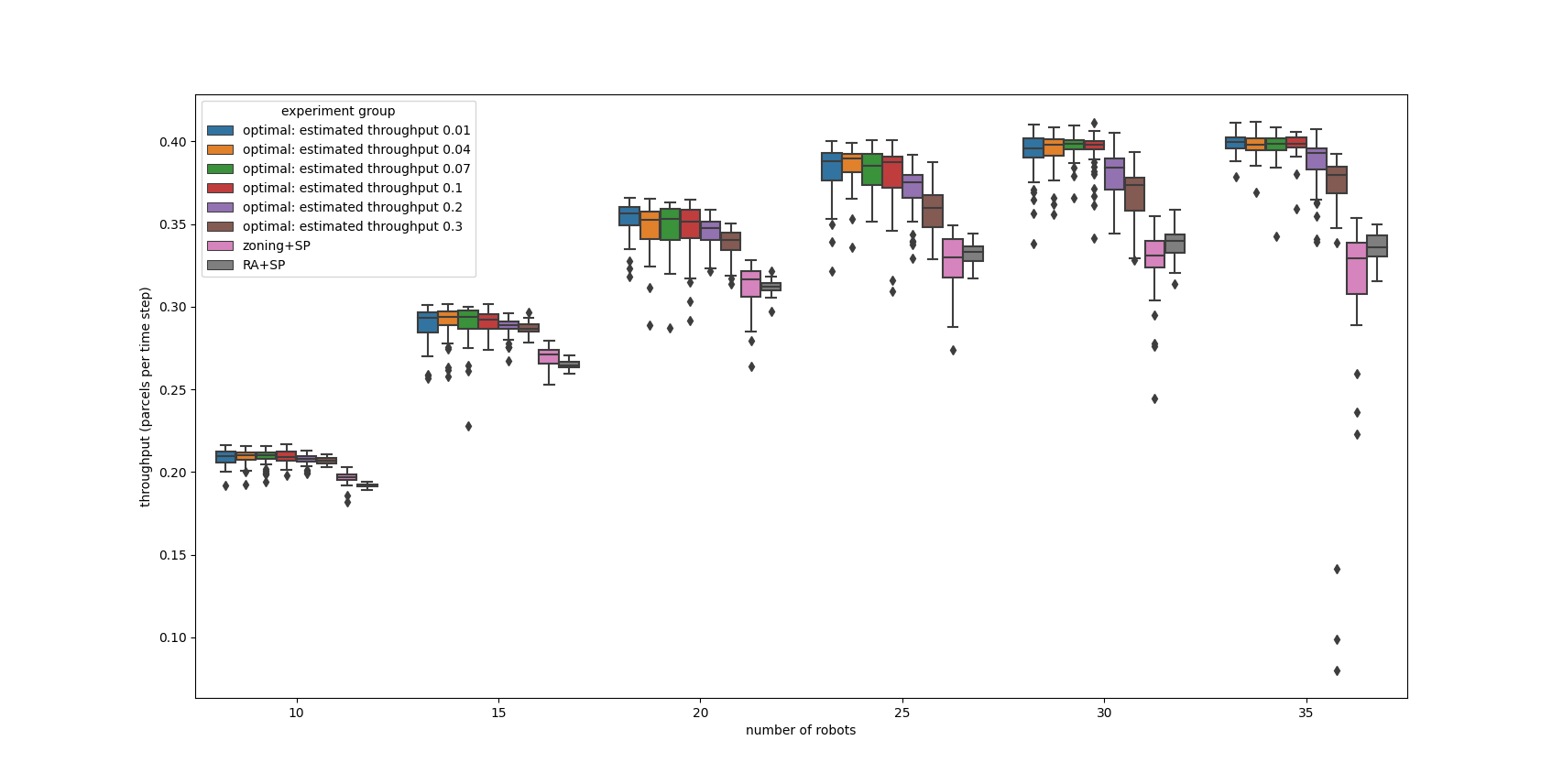}
    \caption{Throughput of Different Methods}
    \label{fig sim results}
\end{figure}

\begin{figure}[ht]
    \centering
    \includegraphics[width=16cm]{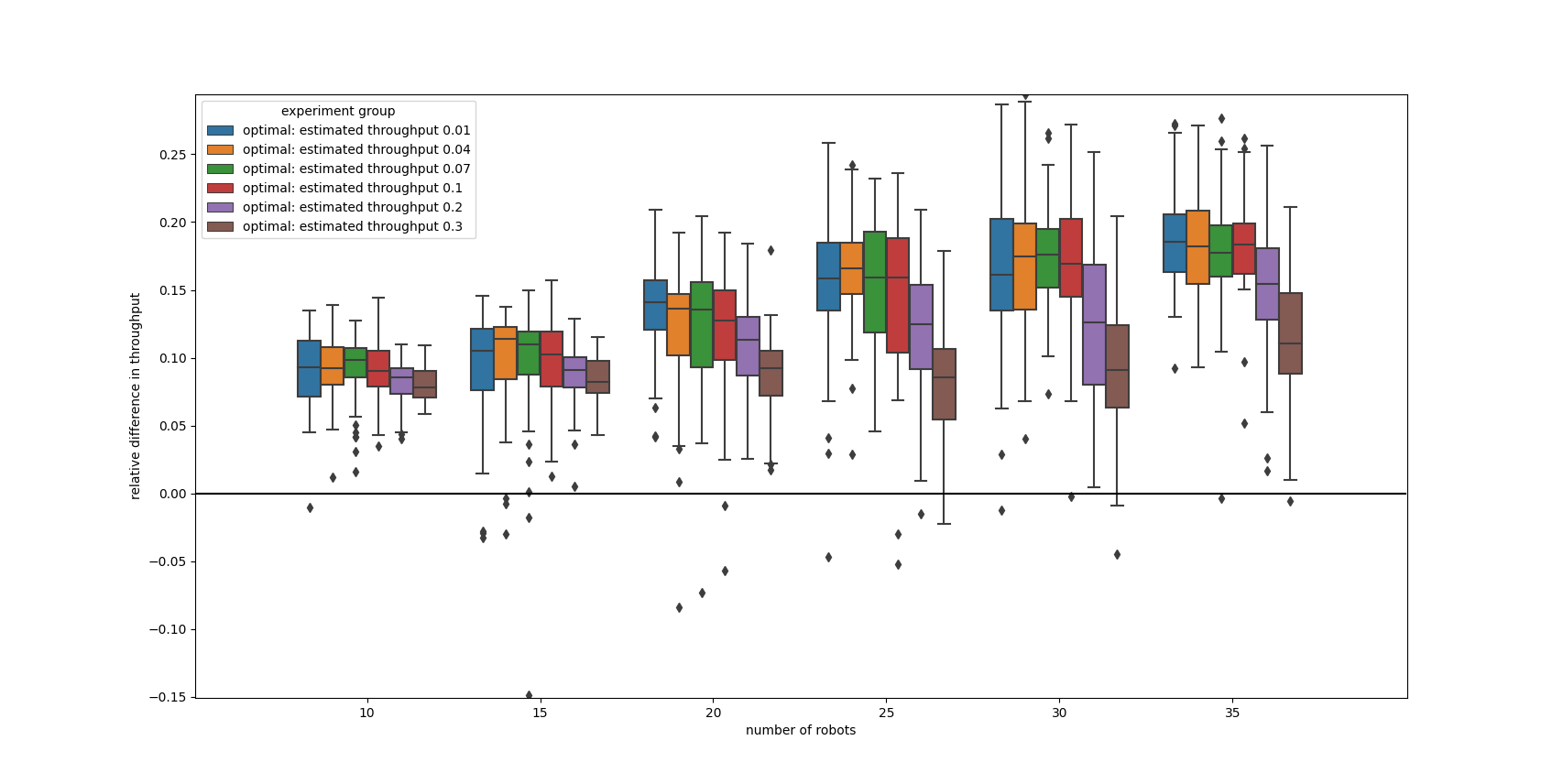}
    \caption{Relative Improvement of Different methods vs RA + SP}
    \label{fig sim improvement}
\end{figure}

We performed 50 trials for each of the following experimental groups:
\begin{itemize}
    \item Random assignment and shortest path (RA + SP): We randomly assign parcels and robots to workstations with equal probabilities. The path-finding algorithm uses CA*, i.e., finding the shortest path on the time-expansion graph with blocked nodes.
    \item Zoning and shortest path (Zoning + SP): We divide the drop-off points into zones according to their distance from the workstations. Each zone has its own robot fleet (in our case, we divide the robots evenly into two zones). Robots cannot travel from one zone to another. The path-finding algorithm is also CA*.
    \item Optimal with an estimated throughput of $\lambda$: When the initial guess of the system throughput is $\lambda$, we run our method, solve the min-cost flow problem, and use the optimal path-flow distribution to help the system to assign and find paths for robots. We set $\lambda = 0.01$, 0.04, 0.07, 0.1, 0.2, 0.3 (unit: parcels handled per time step)
\end{itemize}
The simulation results are presented in Table~\ref{tab: sim results} (for $\lambda=0.1$, 0.2, 0.3 only) and Figure~\ref{fig sim results}. According to these results, our proposed method outperforms the benchmark methods (Zoning + SP or RA + SP), no matter the initial guess of the throughput. From Figure~\ref{fig sim improvement}, the relative improvement is around 10\%--20\% in the throughput. The relative improvement of one trial is calculated by randomly pairing it with one RA + SP trial.

From Figure~\ref{fig sim improvement}, our method performs better when more robots are in the system. Intuitively, our method guides robots to cooperate with each other to avoid congestion and achieve a system optimum instead of greedily seeking the shortest paths. So, when there are more robots, the difference between the system optimum and the individual shortest path solution increases, and our algorithm performs better.

We can see the throughput increase when more robots are in the system. However, the system becomes more unstable since a larger number of robots leads to a greater probability of unsolvable deadlocks. At a certain point, the system becomes saturated. As shown in Figure~\ref{fig sim results}, the average throughput ceases to increase, and there are more outliers when we have more than 25 robots in the system. Each outlier indicates that there is an unsolvable deadlock in the simulation process. The occurrence of deadlocks is highly unpredictable, and an analysis of the deadlock mechanism is beyond the scope of this paper; we treat them as outliers in our experiments.

Zoning does not help much in this experiment because the system size is small. We can see that zoning is helpful when there are 10 robots by reducing the travel distance required. However, when there are more robots, the major delay is caused by congestion instead of free-flow travel, and zoning improvement becomes trivial.
 
\begin{center}
\begin{longtable}{llllllll}
\caption{Simulation Results} \label{tab: sim results} \\
\hline 
        Experiment group & $R$ & Max & q75 & Median & q25 & Min & Mean \\ \hline
        Optimal: estimated throughput 0.1 & 10 & 0.217 & 0.212 & 0.209 & 0.207 & 0.198 & 0.209 \\ 
        Optimal: estimated throughput 0.2 & 10 & 0.213 & 0.209 & 0.208 & 0.206 & 0.199 & 0.207 \\ 
        Optimal: estimated throughput 0.3 & 10 & 0.211 & 0.209 & 0.207 & 0.205 & 0.203 & 0.207 \\ 
        Zoning + SP & 10 & 0.203 & 0.199 & 0.197 & 0.195 & 0.182 & 0.196 \\ 
        RA + SP & 10 & 0.194 & 0.193 & 0.192 & 0.191 & 0.189 & 0.192 \\ 
        Optimal: estimated throughput 0.1 & 15 & 0.301 & 0.296 & 0.292 & 0.287 & 0.274 & 0.291 \\ 
        Optimal: estimated throughput 0.2 & 15 & 0.296 & 0.291 & 0.289 & 0.286 & 0.267 & 0.288 \\ 
        Optimal: estimated throughput 0.3 & 15 & 0.297 & 0.290 & 0.286 & 0.285 & 0.278 & 0.287 \\ 
        Zoning + SP & 15 & 0.280 & 0.274 & 0.271 & 0.265 & 0.253 & 0.269 \\ 
        RA + SP & 15 & 0.271 & 0.267 & 0.265 & 0.263 & 0.259 & 0.265 \\ 
        Optimal: estimated throughput 0.1 & 20 & 0.365 & 0.359 & 0.352 & 0.341 & 0.292 & 0.348 \\ 
        Optimal: estimated throughput 0.2 & 20 & 0.359 & 0.352 & 0.348 & 0.340 & 0.321 & 0.345 \\ 
        Optimal: estimated throughput 0.3 & 20 & 0.350 & 0.345 & 0.340 & 0.334 & 0.314 & 0.338 \\ 
        Zoning + SP & 20 & 0.328 & 0.321 & 0.317 & 0.306 & 0.264 & 0.311 \\ 
        RA + SP & 20 & 0.322 & 0.314 & 0.312 & 0.310 & 0.297 & 0.312 \\ 
        Optimal: estimated throughput 0.1 & 25 & 0.401 & 0.391 & 0.388 & 0.372 & 0.309 & 0.379 \\ 
        Optimal: estimated throughput 0.2 & 25 & 0.392 & 0.379 & 0.375 & 0.366 & 0.329 & 0.371 \\ 
        Optimal: estimated throughput 0.3 & 25 & 0.387 & 0.368 & 0.360 & 0.348 & 0.329 & 0.358 \\ 
        Zoning + SP & 25 & 0.349 & 0.341 & 0.330 & 0.317 & 0.274 & 0.327 \\ 
        RA + SP & 25 & 0.344 & 0.337 & 0.333 & 0.328 & 0.317 & 0.332 \\ 
        Optimal: estimated throughput 0.1 & 30 & 0.411 & 0.400 & 0.398 & 0.395 & 0.342 & 0.395 \\ 
        Optimal: estimated throughput 0.2 & 30 & 0.405 & 0.389 & 0.384 & 0.371 & 0.344 & 0.380 \\ 
        Optimal: estimated throughput 0.3 & 30 & 0.394 & 0.378 & 0.373 & 0.358 & 0.328 & 0.368 \\ 
        Zoning + SP & 30 & 0.355 & 0.340 & 0.331 & 0.324 & 0.244 & 0.325 \\ 
        RA + SP & 30 & 0.359 & 0.344 & 0.340 & 0.332 & 0.314 & 0.338 \\ 
        Optimal: estimated throughput 0.1 & 35 & 0.406 & 0.403 & 0.399 & 0.396 & 0.359 & 0.398 \\ 
        Optimal: estimated throughput 0.2 & 35 & 0.408 & 0.396 & 0.393 & 0.383 & 0.339 & 0.387 \\ 
        Optimal: estimated throughput 0.3 & 35 & 0.392 & 0.385 & 0.380 & 0.368 & 0.080 & 0.360 \\ 
        Zoning + SP & 35 & 0.354 & 0.339 & 0.329 & 0.308 & 0.223 & 0.317 \\ 
        RA + SP & 35 & 0.350 & 0.343 & 0.336 & 0.330 & 0.316 & 0.336 \\ \hline
\end{longtable}
\end{center}

\subsection{Robustness to Initial Throughput Estimation}
\label{s:5.2}
As mentioned in the last paragraph of Section~\ref{s:1} and in the text following Equation~\eqref{eq demand backward}, we need to estimate the ``demand'' for each drop-off point before running the algorithm. This can be done by running simulations or by regression from historical data. There is no way of knowing the actual throughput without running the system. Also, our proposed algorithm does not guarantee an optimal global minimum since the objective is non-convex. In addition, with so many unrealistic assumptions and approximations (namely, the assumption of Poisson flow, the assumption of equal priority in Section~\ref{s:4.1}, and the use of independent service time to replace actual service time in Section~\ref{s:4.1.2}), it seems that there could be significant errors in our so-called optimal flow.

Fortunately, according to our experiment, our method is robust under throughput estimation error. From Figure~\ref{fig sim results}, setting the initial throughput to be 0.01, 0.04, 0.07, 0.1 or 0.2 makes almost no difference. Note that the actual throughput ranges from 0.2 to 0.4 (in units of parcels per time step). A smaller estimation seems to give a better average throughput and stability result. One possible reason is the difference in the optimality gap. Under light traffic, our link cost function \eqref{eq Sij approx}, \eqref{eq cSWk} are more ``linear,'' and the algorithm can give high-quality solutions. When our initial guess of the throughput is large, the objective function becomes more ``nonlinear,'' since the flow variable must be large to satisfy the node demand constraints. Our objective function is nonconvex, and the Frank--Wolfe algorithm does not guarantee an optimal solution; with an approximately linear objective function, the optimality gap can be smaller, so the resulting flow is closer to optimal. Another reason is that our estimation error is in $O(v_{max} R^2)$ according to Theorem \ref{theorem Sij error}, so the approximation error is larger with a larger initial throughput estimation.

\begin{figure}[ht]
\centering
\begin{subfigure}{0.4\textwidth}
    \includegraphics[width=\textwidth]{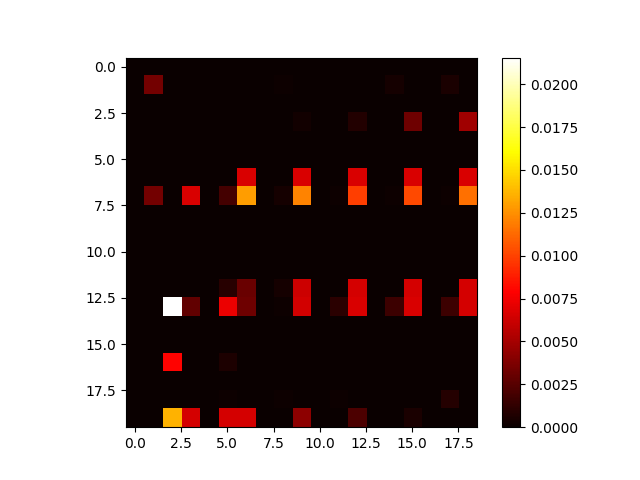}
    \caption[size=9]{Optimal (normalized)}
    \label{fig: math optimal}
\end{subfigure}
\hfill
\begin{subfigure}{0.4\textwidth}
    \includegraphics[width=\textwidth]{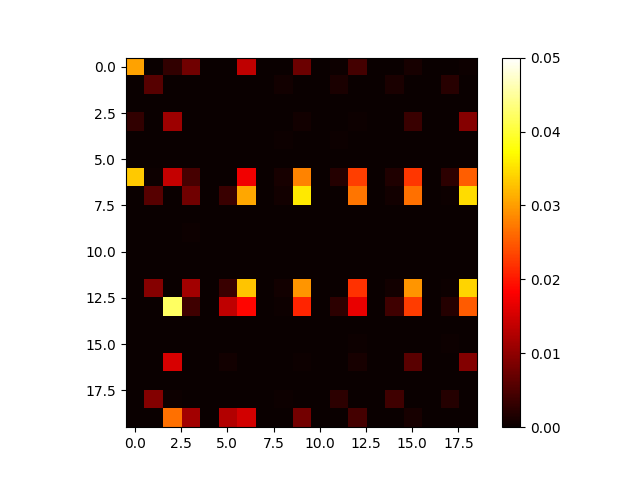}
    \caption[size=9]{Simulated using our method}
    \label{fig: sim optimal}
\end{subfigure}
\hfill
\begin{subfigure}{0.4\textwidth}
    \includegraphics[width=\textwidth]{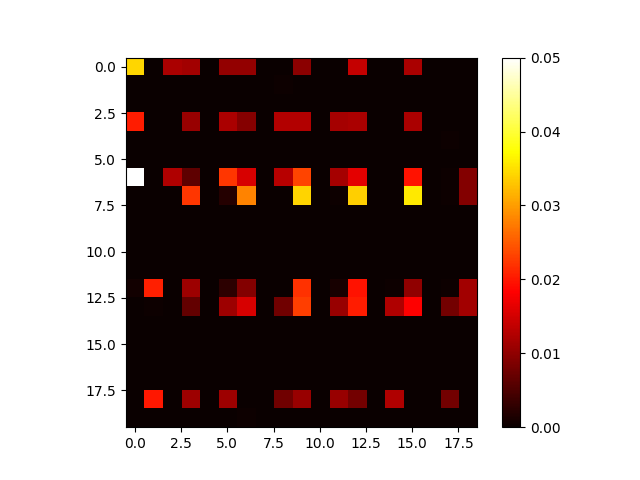}
    \caption[size=9]{Simulated using zoning + SP}
    \label{fig: sim zoning}
\end{subfigure}
    \hfill
\begin{subfigure}{0.4\textwidth}
    \includegraphics[width=\textwidth]{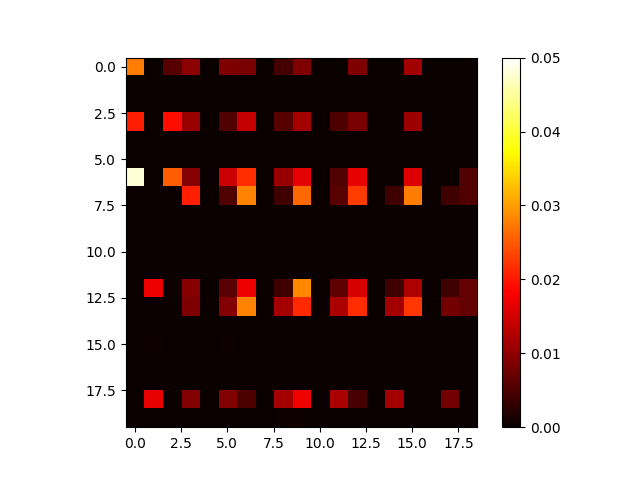}
    \caption[size=9]{Simulated using RA + SP}
    \label{fig: sim RASP}
\end{subfigure}
\caption{Distribution of Turning Flows}
\label{fig: turning flows}
\end{figure}

To get insight into why our method is robust, we show the turning flow distribution in the system in Figure~\ref{fig: math optimal} after running simulations with 20 robots and an initial estimation of throughput of 0.2. The actual flow distribution obtained from the simulation differs greatly from the ``optimal flow'' given by the Frank--Wolfe algorithm because the throughput estimation is different. However, the normalized flow distributions are similar (see Figure~\ref{fig: math optimal} vs. Figure~\ref{fig: sim optimal}), since robots will follow the split proportion of the optimal flow given by our algorithm. We can see lots of robots making turns at cell $(2,12)$ in both the simulated and calculated flow distributions (the small white square in Figure~\ref{fig: math optimal}). Under the guidance of the optimal flow distribution, we can see robots using our method making fewer turns on cell $(0,6)$, which is near the entrance of one workstation. They generally make fewer turns using the leftmost areas, where the workstations are located. Therefore, the robots work together and prevent congestion under our method. Although the calculated flow can be different from the theoretical optimal flow, the calculated flow distribution can still guide the robots to prevent turning near areas that may lead to more congestion, so our method performs better. This method is robust even if we cannot obtain the accurate optimal flow distribution.

\section{Conclusions and Future Work}
We have developed an integrated assignment and decentralized routing method for RSSs. Given the layout, robot fleet, and estimated demand, our method can efficiently assign paths to robots in real-time. In the offline part of the algorithm, we use a directed graph representation of the system and try to find the approximated static-state system-optimal flow distribution. In the online part, we use the given flow distribution to guide all the robots. Owing to our special graph structure, the assignment problem is solved simultaneously when  the path flow is obtained. The online part of the algorithm has a complexity  $O(|\mathcal{V}|)$, and $|\mathcal{V}|$ is proportional to the number of cells in the system, so it can be used in large systems.

Using simulations, we have shown that our method can perform better than zoning or random assignment with prioritized shortest-path routing. We have also shown that our method is robust to uncertainties in throughput estimation. Although we have made many assumptions when deriving the min-cost problem, the resulting flow can still guide robots to achieve better performance by avoiding turning around easy-to-congest areas.

There are several directions for future studies. For one thing, we still do not fully understand the mechanism of saturation, which leads to outliers in Figure~\ref{fig sim results}. Moreover, our current method is not dynamic. Incorporating real-time traffic state information could potentially improve the assignment and routing results. In addition, our model can be easily extended to similar systems such as RMFSs, and more real-world industry-level experiments are needed to show the full potential and limits of our method.


%
\begin{APPENDIX}{Table of Notations}

\begin{center}
\begin{longtable}{ll}
\caption{Notations} \label{tab: notation} \\
\hline 
        Symbol & Meaning \\ 
        \hline
        Set and elements\\
        \hline
        $\{C_1,...,C_N\}$ & set of cells\\
        $\{D_1,...,D_{n_D}\}$ & set of dropoff points\\
        $\{W_1,...,W_{n_W}\}$ & set of workstations\\
        $\mathcal{G}=(\mathcal{V},\mathcal{A})$ & graph representation of the system\\
        $\mathcal{R}_F$, $\mathcal{R}_B$ & set of forward and backward paths\\
        $S$ & source node\\
        $T$ & sink node\\
        \hline
        Parameters\\
        \hline
        $N$ & number of cells\\
        $n_W$ & number of workstations\\
        $n_D$ & number of dropoff points\\
        $R$ & number of robots\\
        $T_1$ & time to move from one cell to its neighboring cell\\
        $T_2$ & time to take a 90-degree turn\\
        $T_{drop}$ & time to drop an item, can be random\\
        $T_{load}$ & time to load an item, can be random\\
        $d_{D_k}$ & estimated dropping demand of dropoff point $D_k$\\
        \hline
        Variables\\
        \hline
        $B_j$ & random service time for a robot on cell $j$\\
        $c_{ij}$& cost on arc $(i,j)$\\
        $f_r^F, f_r^B, f_r$& forward/backward flow intensity on path $r$\\
        $G_j$ & service time on cell $j$ if no downstream cells are blocked\\
        $I_{k,j}$ & $=1$ if there is an robot on $k$ heading to $j$\\
        $R_j$ & unfinished working time on cell $j$\\
        $RC_r$ & cost of path $r$\\
        $S_{ij}$ & random waiting time for robot entering cell $j$ from $i$\\
        $v_{ij}$ & arc flow of robots on $(i,j)$\\
        $v_k$ & total flow that come through cell $k$\\
        $v_{max}$ & max cell flow $:=\max_{k} v_k$\\
        $v_k^{(1)}, v_k^{(2)},v_k^{(3)}$ & through flow, turning flow, and dropping-off flow on cell $k$\\
        \hline
\end{longtable}
\end{center}

\end{APPENDIX}
%
%


\bibliographystyle{informs2014trsc} 
\bibliography{reference.bib} 

\end{document}